\documentclass[letterpaper, 10 pt, conference]{ieeeconf}  

\IEEEoverridecommandlockouts                              

\usepackage{amsmath} 
\usepackage{graphicx}
\usepackage{url}
\usepackage{breakurl}
\usepackage[breaklinks]{hyperref}

\DeclareMathOperator{\dis}{d}
\title{\LARGE \bf
Deep Reinforcement Learning-Based Mapless Crowd Navigation with Perceived Risk of the Moving Crowd for Mobile Robots
}

\author{Hafiq Anas, Wee Hong Ong, and Owais Ahmed Malik
\thanks{All authors are with School of Digital Science, Universiti Brunei Darussalam,
        Jalan Tungku Link, Brunei.
        {\tt\small hafiq.anas@gmail.com, [weehong.ong, owais.malik]@ubd.edu.bn}}%
}

\begin{document}

\maketitle
\thispagestyle{empty}
\pagestyle{empty}

\begin{abstract}

Current state-of-the-art crowd navigation approaches are mainly deep reinforcement learning (DRL)-based. However, DRL-based methods suffer from the issues of generalization and scalability. To overcome these challenges, we propose a method that includes a Collision Probability (CP) in the observation space to give the robot a sense of the level of danger of the moving crowd to help the robot navigate safely through crowds with unseen behaviors. We studied the effects of changing the number of moving obstacles to pay attention during navigation. During training, we generated local waypoints to increase the reward density and improve the learning efficiency of the system. Our approach was developed using deep reinforcement learning (DRL) and trained using the Gazebo simulator in a non-cooperative crowd environment with obstacles moving at randomized speeds and directions. We then evaluated our model on four different crowd-behavior scenarios. The results show that our method achieved a 100\% success rate in all test settings. We compared our approach with a current state-of-the-art DRL-based approach, and our approach has performed significantly better, especially in terms of social safety. Importantly, our method can navigate in different crowd behaviors and requires no fine-tuning after being trained once. We further demonstrated the crowd navigation capability of our model in real-world tests.

\end{abstract}

\section{INTRODUCTION}

Autonomous mobile robots are increasingly being deployed in human living spaces to provide services such as servicing, delivering, and guiding. In these situations, robots must navigate crowded environments with moving humans at varying speeds. This is known as crowd navigation, or social navigation in some cases. The classical navigation approaches of using global and local planners struggle in highly dense crowded environments and would often result in the robot being stuck in an endless replanning state. To address the limitations of classical planning approaches in crowded environments, recent research works have focused on deep reinforcement learning (DRL) methods \cite{9409758}. Many recent DRL-based works are mapless and have empirical evidence that demonstrates the capability of DRL-based approaches with 2D laser scans for crowd navigation \cite{9197148}\cite{8202134}\cite{zheloo2018curiosity}\cite{8461113}. However, autonomous mobile robots currently deployed in real-world applications remain dependent on cooperation from the crowd (humans) during navigation.

In this paper, we propose a DRL-based approach using 2D laser scans for local navigation in crowded environments. Our approach incorporates risk perception of the moving crowd into the observation space of our DRL system. Particularly, we utilized the collision probability to identify $K$ most dangerous obstacles to pay attention in the observation space of the robot. The inclusion of risk perception in the observation space allows the robot to more precisely assess the risk during crowd navigation even with an unseen crowd behavior. By limiting the attention to a small number of moving obstacles, the system can scale to high crowd density. In addition, we included local waypoints into the reward function to improve the efficiency in learning to reach the target goal. We evaluated our approach in different crowd behavior settings and compared our results with the results of a recent state-of-the-art DRL-based approach \cite{9197148} under the same set of test conditions. The main contributions of this work are the ideas of including risk perception of $K$ most dangerous obtacles in the observation space to allow the robot to perceive the danger level of the moving crowds to improve local path planning, and the use of waypoints to improve global path planning. We have also verified the ideas through successful implementations in both simulation and real-world settings.

\section{RELATED WORKS}

An early DRL-based solution to address the crowd navigation problem was the CrowdMove implementation \cite{8461113}. CrowdMove was trained and tested in multiple dynamic environments using commonly used observation states, such as the robot's own velocity and relative target goal position. The authors concluded that their robot was able to avoid moving obstacles in real-world tests and that their trained model could be generalized to different environment settings unseen during training. We note that their approach relied on providing sufficient variation in the training data of multiple dynamic environments to improve generalization. 

In a recent work, Jin et al. \cite{9197148} proposed that a robot moving in a crowded environment should have human-awareness competencies. Therefore, they implemented this through their reward setup by incorporating two conditions: ego-safety and social-safety violations. Using this reward setting, they trained their robot in one crowded environment and tested it in four different crowd behavior environments with varying number of moving obstacles. They achieved significant performance improvement over the then state-of-the-art DRL-based crowd navigation method, CADRL \cite{7989037}. We consider their work a representative current state-of-the-art in crowd navigation using 2D laser scans. Jin et al. \cite{9197148} used ego and social scores in their reward function to model human-awareness, but this approach is limited by the lack of access to such information during deployment. In this sense, their model will require a large amount of training to infer the perceived risk from the typical observation states in different scenarios. Unlike Jin et al. \cite{9197148}, we incorporate perceived risk or human-awareness into the observation space, which allows the robot to perceive potential risk during testing or deployment.

\section{APPROACH}

\subsection{Problem Formation}

\begin{figure*}
	\centering
	\includegraphics[width=0.8\textwidth]{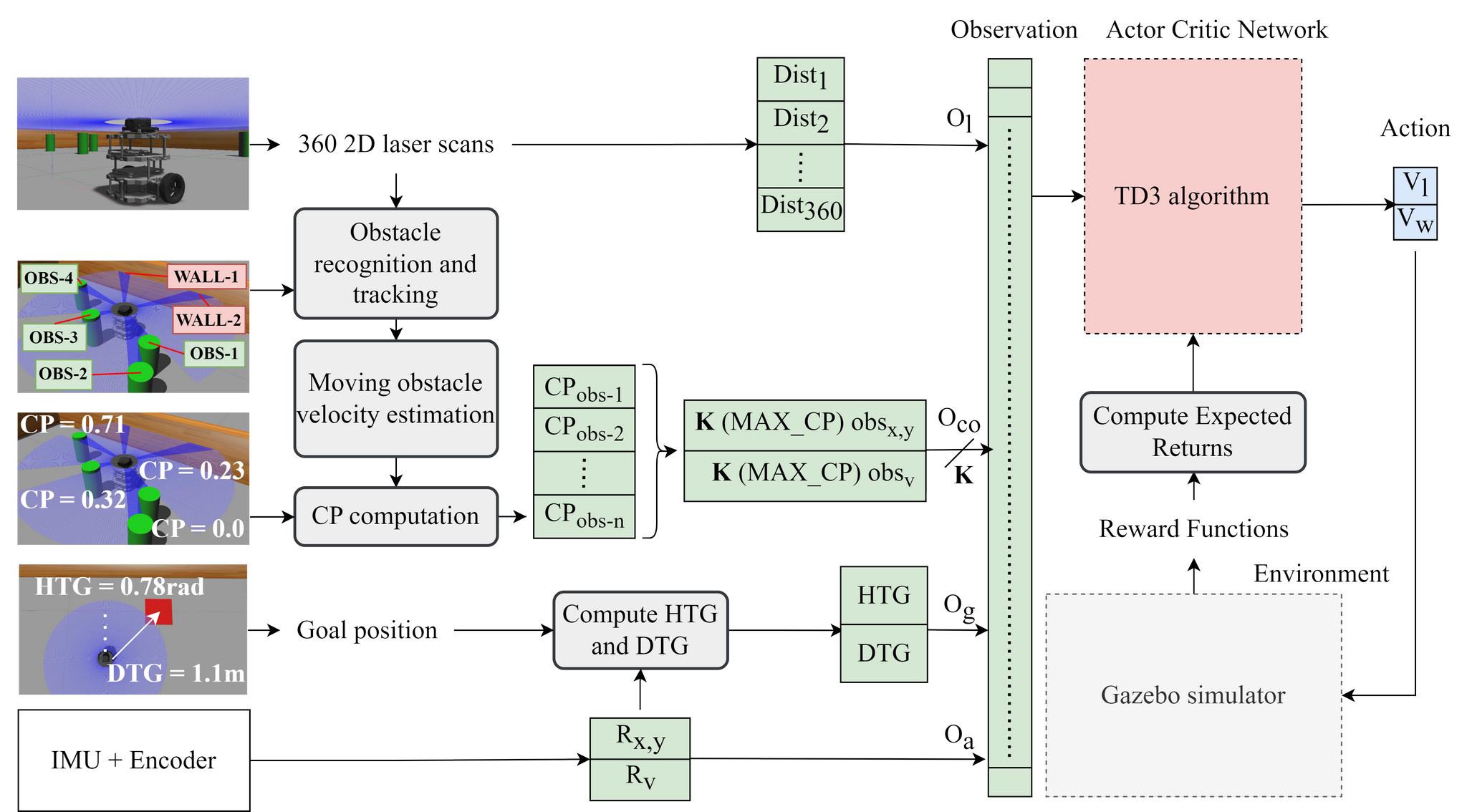}
	\caption{Deep Reinforcement Learning system structure.}
	\label{fig:drl_system_overview}
\end{figure*}

Our proposed method builds upon the techniques used in the existing DRL-based methods, with the addition of perceived risk in the observation states and prioritizing the $K$ most dangerous obstacles within a crowd. To determine which obstacles to prioritize, we compute the collision probability of all tracked moving obstacles within the robot's field of view (FOV) and focus on the top $K$ obstacles with the highest probability of collision where $K$ corresponds to the number of moving obstacles for the robot to pay attention to during navigation. Fig. \ref{fig:drl_system_overview} shows the overview of our deep reinforcement learning (DRL) system. The components of the system are described in the following subsections.

\subsubsection{\textbf{Observation space}}

The observation space contains input features to learn and perform crowd navigation behavior that solves both local and global navigation. To solve the global navigation problem, the information of relative distance to goal (DTG) and orientation (heading to goal, HTG) of the target goal location is used as the goal-related observations \textit{o}\textsubscript{\textit{g}}. Meanwhile, \textit{o}\textsubscript{\textit{l}} contains distance information from the 2D laser scan sensor that describes the static environment of the robot and is used to solve the local navigation problem. Given that a crowded environment is associated with moving obstacles, we added agent-related observations \textit{o}\textsubscript{\textit{a}} and critical obstacle observation \textit{o}\textsubscript{\textit{co}} to the observation space. \textit{o}\textsubscript{\textit{a}} contains the robot's position (\textit{R}\textsubscript{\textit{x,y}}) and velocity (\textit{R}\textsubscript{\textit{v}}) estimated from its encoder and inertia sensor. \textit{o}\textsubscript{\textit{co}} describes the position (\textit{obs}\textsubscript{\textit{x,y}}) and velocity (\textit{obs}\textsubscript{\textit{v}}) of the $K$ most dangerous moving obstacles (critical obstacles). We define an observation as \textit{o} = [\textit{o}\textsubscript{\textit{l}}, \textit{o}\textsubscript{\textit{g}}, \textit{o}\textsubscript{\textit{a}}, \textit{o}\textsubscript{\textit{co}}] which describes the partial environment the robot can observe at a given time.

\begin{figure}
	\centering
	\includegraphics[width=0.5\textwidth]{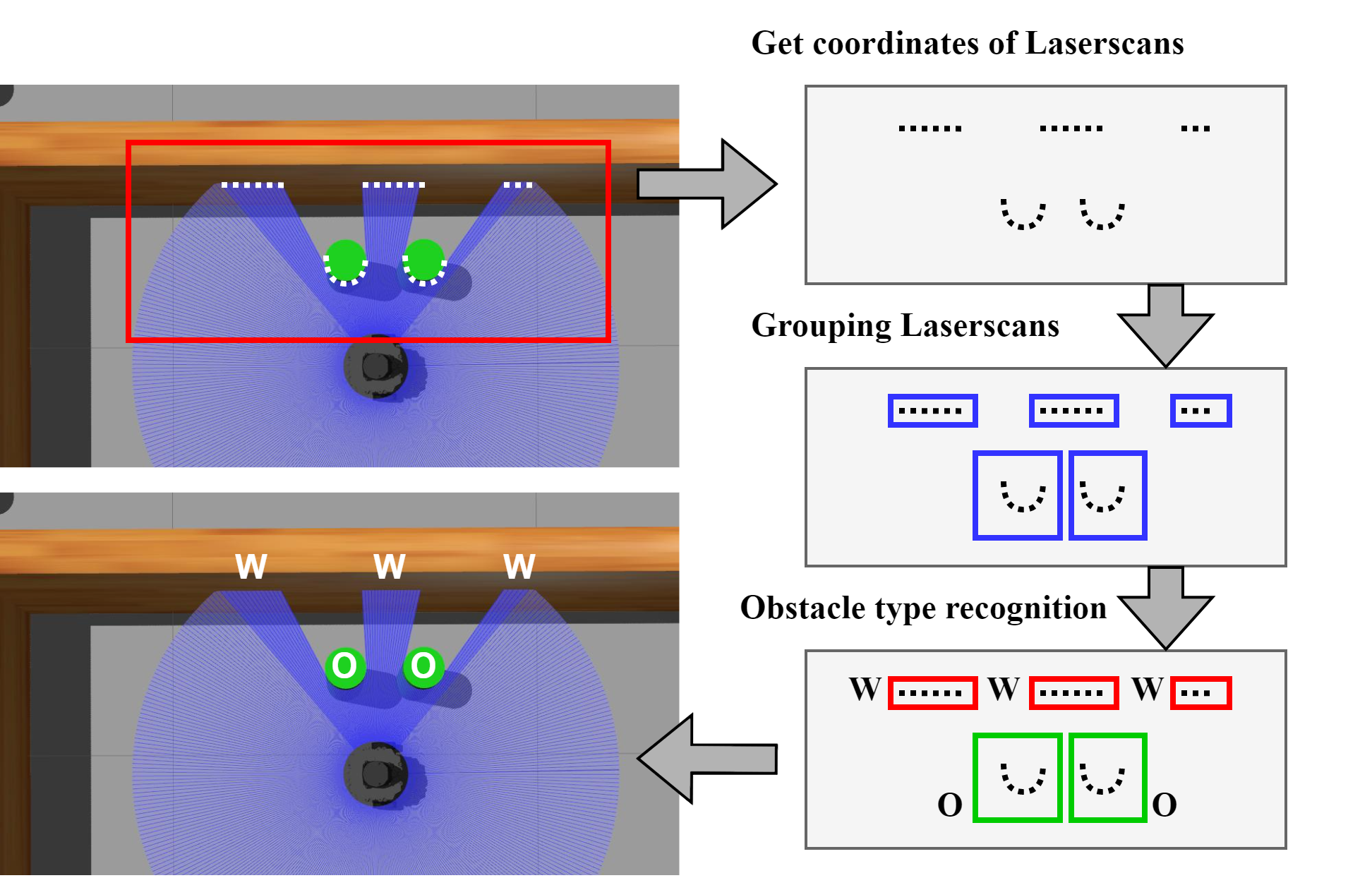}
	\caption{Wall (W) and Moving obstacle (O) recognition process.}
	\label{fig:object_detection}
\end{figure}

Obstacles tracking was implemented for obstacle velocity estimation and computation of Collision Probability (CP) from the 2D laser scans \textit{o}\textsubscript{\textit{l}}. Using the 2D laser scans \textit{o}\textsubscript{\textit{l}}, the robot can differentiate between the wall and the moving obstacles, and hence moving obstacles can be tracked. The method is illustrated in Fig. \ref{fig:object_detection}. First, the 2D laser scans' values in \textit{o}\textsubscript{\textit{l}} are converted to cartesian coordinates using the robot's position and orientation. We use Kuhn-Munkres algorithm \cite{kuhn1955hungarian}\cite{munkres1957algorithms} to segment the coordinate values of the scans into $N$ number of groups that correspond to the possible number of obstacles seen by the robot. Then for each group, we compute the gradient of each pair of laser scans' coordinates to determine the object type (wall or obstacle). A scan group is determined to be a wall object if all computed gradients are close to zero. While an obstacle type is confirmed otherwise. Finally, we separate the scans that belong to different object types and use the center scan of each group as the position of the object for tracking, velocity estimation, and CP computation.

We define the Collision Probability (CP) as the sum of two component probabilities: the probability of collision based on the time to collision ($P_{c-ttc}$) and the probability of collision based on the distance to the obstacle ($P_{c-dto}$).  We argue that the addition of distance to obstacle (\textit{dto}) information allows the robot to better perceive the collision probability with a moving obstacle in the crowd. For example, an obstacle moving slowly near the robot can still pose substantial risk of collision while an obstacle moving fast toward the robot from a far distance is less dangerous. Therefore, a balance between the two CP components is made as given in (\ref{eqn:cp_total}).

\begin{equation}
	\label{eqn:cp_total}
	CP = \alpha \cdot P_{c-ttc} + (1 - \alpha) \cdot P_{c-dto}
\end{equation}
where $\alpha$ $\in$ [0, 1] is the parameter that decides the weight of collision probabilities $P_{c-dto}$ and $P_{c-ttc}$. We have set $\alpha = 0.5$ in the experiments reported in this paper.

\begin{figure}
	\centering
	\includegraphics[width=0.35\textwidth]{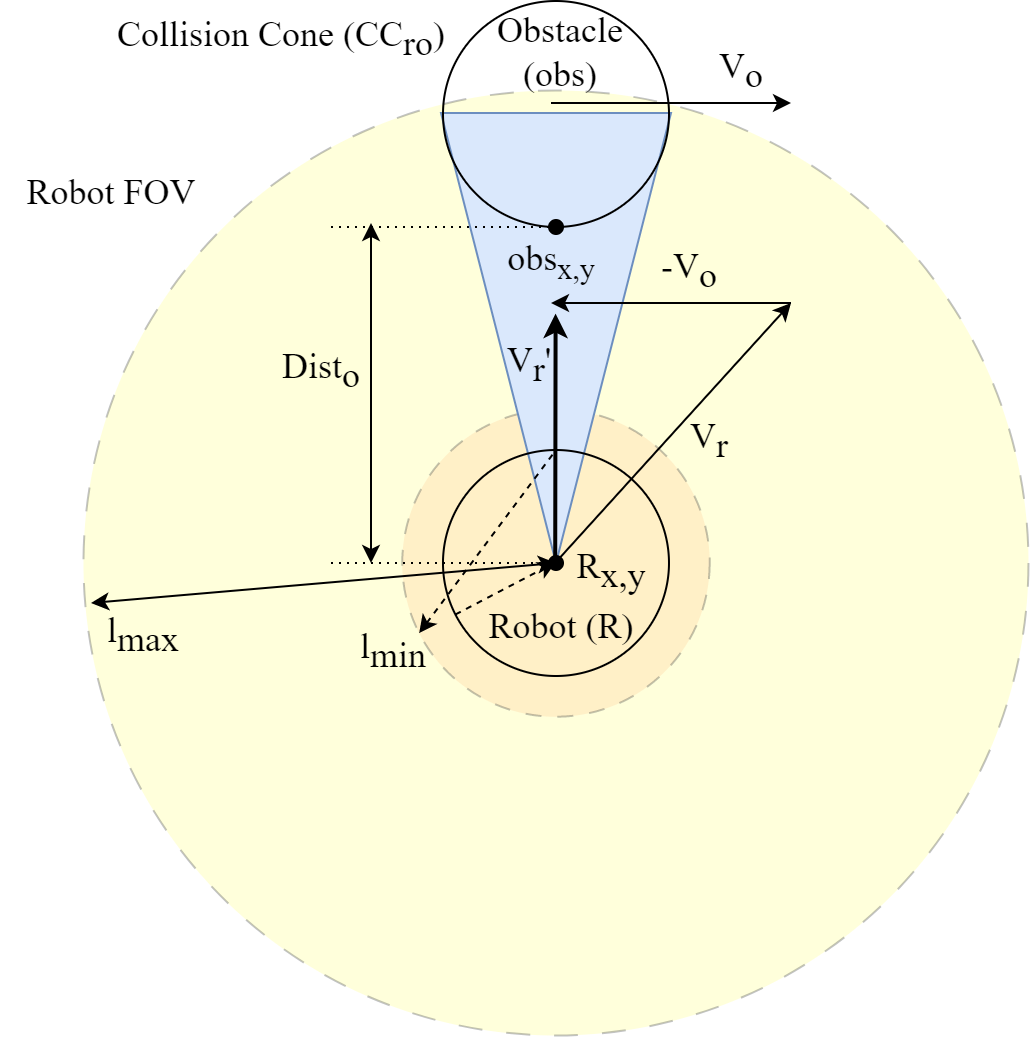}
	\caption{The Collision Cone and related information.}
	\label{fig:cp_estimation}
\end{figure}

The calculation of collision probabilities uses the Collision Cone (CC) concept in \cite{8789446}\cite{fiorini1998motion}. Fig. \ref{fig:cp_estimation} shows an illustration of the CC and the related information that are used to estimate the two components of CP. $P_{c-ttc}$ is computed based on time to collision as defined in (\ref{eqn:cp_ttc}). $P_{c-dto}$ is computed based on relative distance to obstacle and defined in (\ref{eqn:cp_dto}).

\begin{equation}
	\label{eqn:cp_ttc}
	P_{c-ttc}= 
	\begin{cases}
	\min(1, \frac{0.15}{t}),& \text{if } V_r'\in CC_{ro}\\
	0              & \text{otherwise}
	\end{cases}
\end{equation}
where $t$ is the time-to-collision (TTC) when the relative velocity $V_r'$ between the robot and the obstacle lies within the Collision Cone $CC_{ro}$. $V_r' = V_r - V_o$ is the resultant velocity between the robot velocity $V_r$ and obstacle velocity $V_o$. $t = Dist_o / V_r'$ is the time to collision. $CC_{ro}$ is the collision cone area between the robot and obstacle. Finally, 0.15 corresponds to the timestep value of the robot in seconds for executing its velocity commands.

\begin{equation}
	\label{eqn:cp_dto}
	P_{c-dto}= 
	\begin{cases}
	\frac{l_{max} - Dist_o}{l_{max} - l_{min}},& \text{if } Dist_o < l_{max}\\
	0              & \text{otherwise}
	\end{cases}
\end{equation}
where $l_{max}$ and $l_{min}$ are the maximum and minimum range of the laser scan respectively. $Dist_o$ is the distance from the robot to the obstacle of interest.

CP is computed for each obstacle in the list of tracked obstacles and, the position (\textit{obs}\textsubscript{\textit{x,y}}) and velocity (\textit{obs}\textsubscript{\textit{v}} = \textit{V}\textsubscript{\textit{o}}) of the $K$ obstacles with the highest CP are included in the observation space as $o_{co}$. These $K$ obstacles are seen as most probable to be in collision with the robot.

\subsubsection{\textbf{Action space}}

An action is defined as \textit{a} = [\textit{V}\textsubscript{\textit{l}}, \textit{V}\textsubscript{\textit{w}}] which is sampled from a stochastic policy $\pi$ given observation \textit{o}$\colon$\textit{a} $\sim$ $\pi$(\textit{a} $\mid$ \textit{o}) where \textit{V}\textsubscript{\textit{l}} is the linear velocity within the range [0, 0.22] $ms^{-1}$ and \textit{V}\textsubscript{\textit{w}} is the angular velocity within the range [-2.0, 2.0] $rad.s^{-1}$.

\subsubsection{\textbf{Reward functions}}

The reward function consists of the following terms:

\begin{equation}
R = R_{step} + R_{dtg} + R_{htg} + R_{goal} + R_{col} + R_{wp}
\end{equation}

$R_{step} = -2$ is a negative reward given to the robot for every step and serves to encourage the robot to avoid abusing the $R_{dtg}$ and $R_{htg}$ rewards by oscillating around the goal location without reaching it.

\begin{equation}
R_{dtg}= 
\begin{cases}
+1,& \text{if } {\dis(r, g)}_{t} < {\dis(r, g)}_{t-1}\\
0              & \text{otherwise}
\end{cases}
\end{equation}

$R_{dtg}$ is a positive reward given to the robot whenever the distance from the robot to the target goal location $\dis(r, g)$ has reduced between the current and previous timestep.

\begin{equation}
R_{htg}= 
\begin{cases}
+1,& \text{if } {\theta(r, g)}_{t} < {\theta(r, g)}_{t-1}\\
0              & \text{otherwise}
\end{cases}
\end{equation}

Similarly, $R_{htg}$ is a positive reward given whenever the relative heading $\theta(r, g)$ has decreased.

$R_{goal}=+200$ is a large positive reward given to the robot when it reaches the target goal. If a collision occurs, a penalty $R_{col}=-200$ is given instead. $R_{wp}=+200$ is a large positive reward given to the robot when it reaches a target local waypoint. The next waypoint is computed by finding the point of intersection between the circle boundary from the robot center and a straight line towards the target goal position from the robot. The radius of the circle region is the maximum laser scan distance of 0.6m. The inclusion of $R_{wp}$ addresses the problem of sparse goal reward $R_{goal}$ and improves the policy convergence as well as the quality of the global path planning.

\subsection{Deep Reinforcement Learning}

We use the Twin Delayed Deep Policy Gradient (TD3) \cite{fujimoto2018addressing} algorithm with the default parameters to learn the navigation policy.

\subsubsection{\textbf{Model training}}

The robot was trained in the Gazebo simulator using Robotis TurtleBot3 Burger platform and is equipped with an LDS-01 360-degree 2D laser scanner and XL430-W250 encoder motors. The resolution of the laser scanner is 360 with a minimum and maximum range set to 0.105m and 0.6m respectively. The training process was done once in a 2m x 2m space with walls and random moving obstacles moving at a random speed of up to 0.2m/s in random directions. The moving obstacles were non-cooperative so they will ignore the robot’s presence and can collide with the robot. The model was trained for 3000 episodes with the stopping criteria of collision with an obstacle or having reached the goal.

\subsubsection{\textbf{Model testing}}

The robot was trained in one simulation setting using TD3 and tested in different crowd behavior settings as described in Section \ref{sec:evaluation-section}. The crowd was non-cooperating.

\section{EVALUATION}\label{sec:evaluation-section}

\begin{figure}
	\centering
	\includegraphics[width=0.4\textwidth]{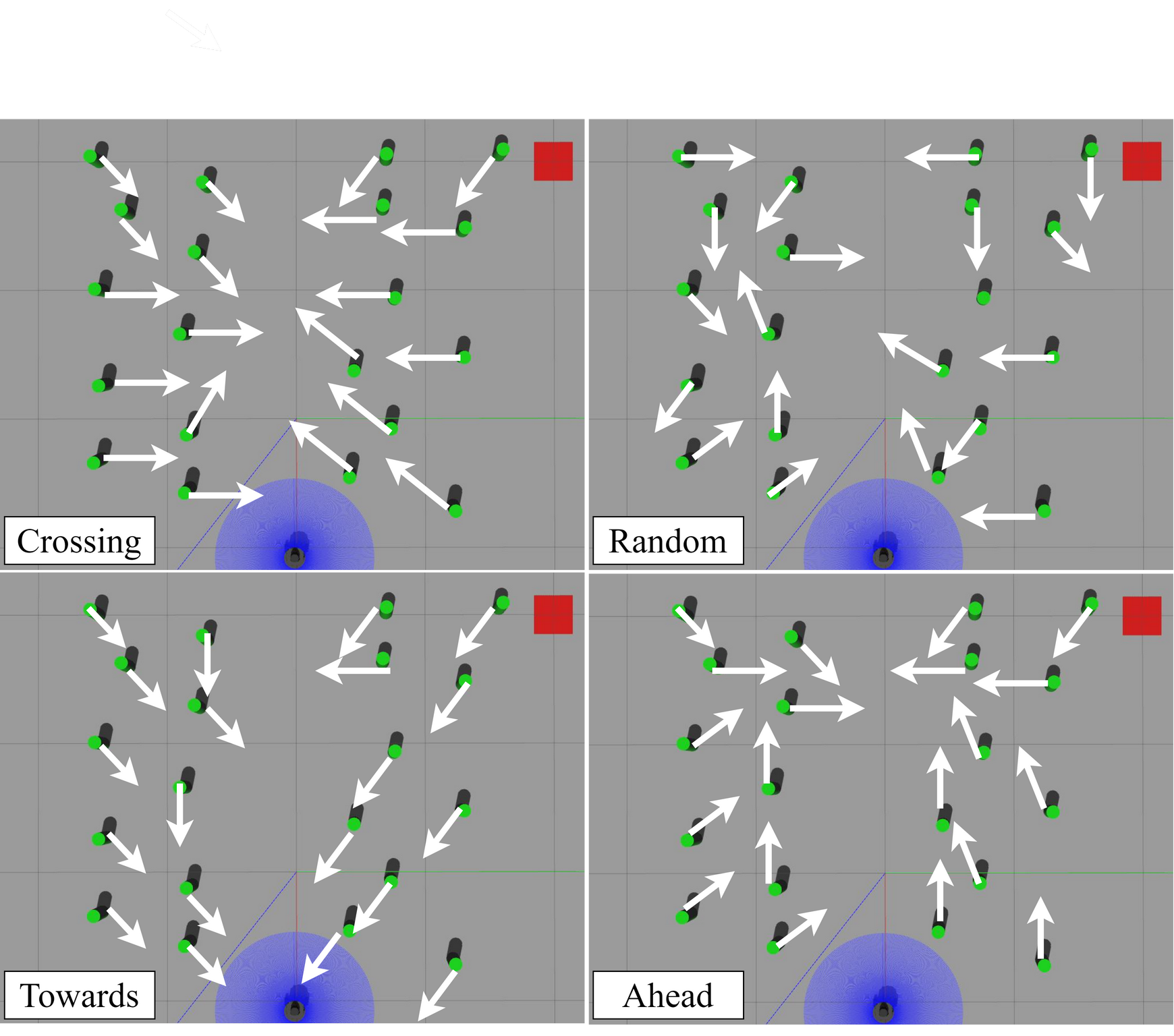}
	\caption{The four crowd behavior settings with 20 obstacles in Gazebo. \textbf{Crossing}: The robot has to navigate through the crowd moving in the crossing directions. \textbf{Towards}: The crowd is moving toward the general direction of the robot. \textbf{Ahead}: The crowd is moving ahead of the robot. \textbf{Random}: The crowd is moving in random directions.}
	\label{fig:testing_environments}
\end{figure}

We evaluated our robot in different crowd behavior environments similar to \cite{9197148}: crossing, towards, ahead, and random. For each crowd behavior, the model was tested with crowd densities of twenty moving obstacles. For each crowd behavior setting, we computed the average of each metric over 10 separate runs. Fig. \ref{fig:testing_environments} shows the four crowd behavior settings with 20 moving obstacles. For the Ego and Social Policy of \cite{9197148}, the authors have tested it on the four crowd behavior settings with twelve moving obstacles.

To quantify performance, we used the same evaluation metrics from Jin et al.’s \cite{9197148} work: success rate (\%), arriving time (s), ego score (0-100) and social score (0-100).  Let $k$ be the number of ego-safety violation steps, and $N$ be the total steps to reach the goal, then $Ego\_Score = (1 - k/N) * 100$. Let $m$ be the number of social-safety violation steps, then $Social\_Score = (1 - m/N) * 100$. An ego-safety violation is determined when an obstacle comes close to the robot within the ego radius of the robot. We have set the ego radius 0.787 times of the largest width of the Turtlebot3 base on the same ratio used in \cite{9197148}. In \cite{9197148}, they have determined the social-safety violation when two rectangular spaces computed from the speed of the robot and the speed of an obstacle intersect. The rectangular spaces are similar to the concept of Collision Cone in our case. For the social-safety violation, we have used the time to collision probability ($P_{c-ttc}$) to determine if our robot is in a collision trajectory course with an obstacle when the $P_{c-ttc}$ value is greater than 0.4. 

For comparison purposes, we have determined by watching Jin et al.'s demonstration video \cite{9197148} that their obstacles were moving about five times slower than the max speed of their robot (1.5m/s). In our case, we performed tests in Gazebo with the $K$ obstacles moving at five times slower than our robot’s max speed (0.22m/s).

\begin{figure}
	\centering
	\includegraphics[width=0.4\textwidth]{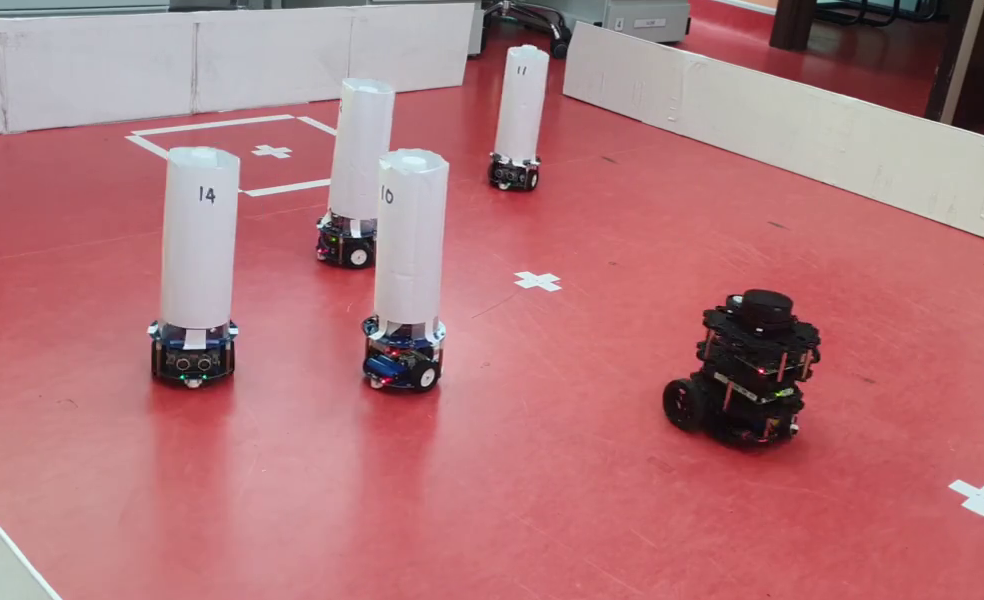}
	\caption{Real world test.}
	\label{fig:real_world_test}
\end{figure}

In real world test, the robot was tested in the same four crowd behaviors shown in Fig. \ref{fig:testing_environments}. We have used mobile robots of similar size to the Turtlebot3 as moving obstacles. The moving obstacles were manually teleoperated by humans. It was difficult to teleoperate the obstacles when there are many of them, hence we have limited the real-world tests to four obstacles. Fig. \ref{fig:real_world_test} shows a photo of the real world test setup. A video recording is available at \cite{hafiqanas2023git},

\section{RESULTS AND DISCUSSION}

\subsection{Crowd Navigation}

\begin{figure*}
	\centering
	\includegraphics[width=\textwidth]{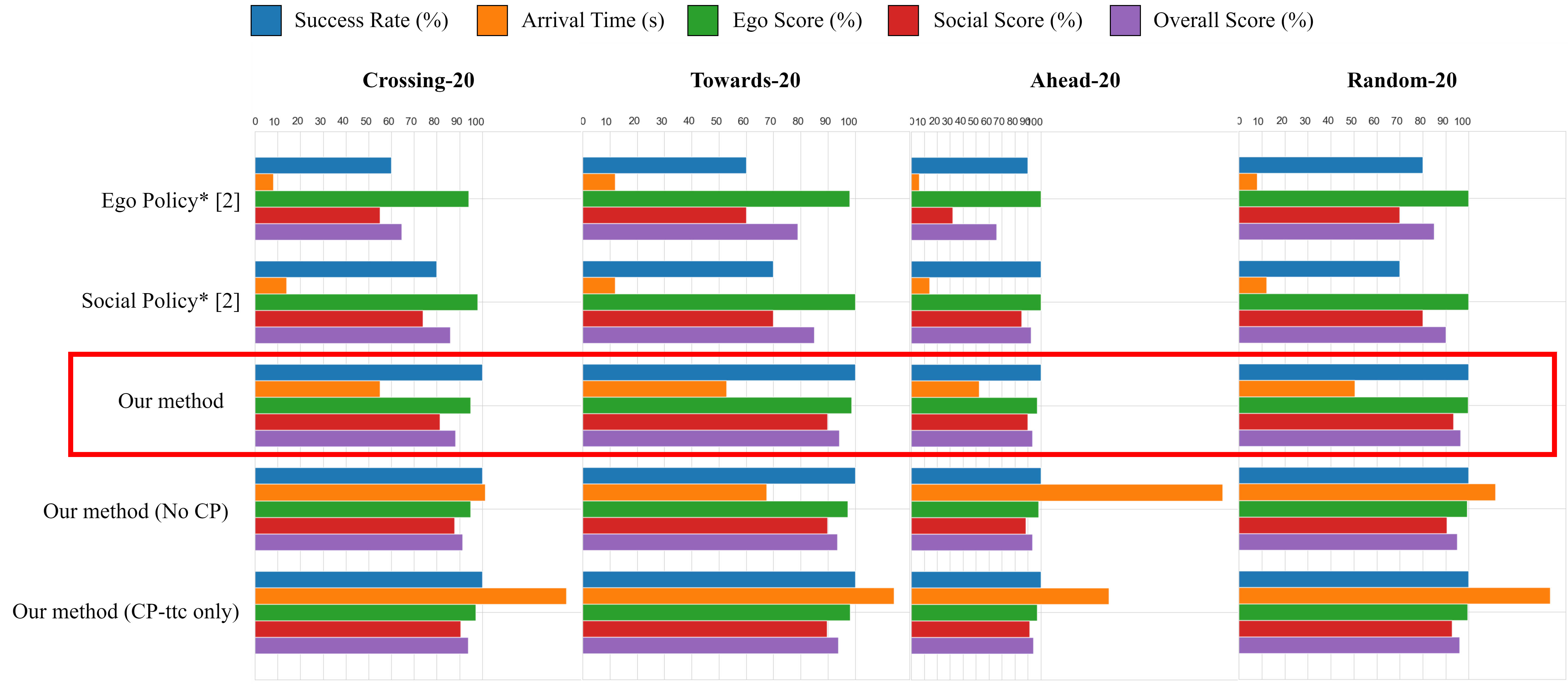}
	\caption{Comparison between our approach and Jin et al.'s results \cite{9197148}. The label of the crowd behavior settings follows the pattern of "behavior-obstacles", e.g., crossing-20 is the test environment with crossing crowd behavior and 20 moving obstacles. Our method here is K = 1. Overall Score is the average of Ego Score and Social Score. * denotes the method was tested with twelve obstacles by the authors. The bottom two rows are the results of the ablation study discussed in Section~\ref{subsection:ablation}}
	\label{fig:results_comparison}
\end{figure*}

Fig. \ref{fig:results_comparison} shows the evaluation results of our method in comparison to the results Jin et al. \cite{9197148}. The navigation policy was learned with $K=1$ and tested with $K=1$. Our robot was able to avoid obstacles with smooth maneuver that resulted in 100\% Success Rates (SR) in all test environments. Jin et al.’s \cite{9197148} success rates were between 60\% to 100\% with only 10 out of 20 tests achieved 100\% success rate. While the arrival times of our robot were longer than the results of \cite{9197148}, we note that their robot was traveling at a speed (1.5m/s) about 7 times faster than our robot (0.22m/s). Taking into account the speed difference, the arrival time of our robot was relatively short.

Looking at our result, we can see that our robot took the least time to navigate in the Ahead-20 environment. This is due to the crowd was moving away from the robot. We have observed that Ego-safety and social-safety violations do not necessarily result in collisions. They are measures of how risky the robot was navigating in the crowd. The Ego Score and Social Score were lowest in Crossing-20 which indicate that Crossing-20 was more risky than the other crowd behaviors. During the course of the project, given that the social scores are inversely proportional to the total number of steps ($N$) in an episode, we have observed that social scores could be biased towards being high despite incurring high social-safety violation counts if the robot spends the majority of its navigation time in free space with no obstacles. We suggest that a more accurate computation of the social score could be made by defining the $N$ as the number of steps where an obstacle is within the detection range of the robot.

We observed that the robot struggled to converge to a good navigation policy when relying only on the sparse goal reward. The robot needed to learn to reach the goal position while also avoiding the moving obstacles. By adding waypoints reward, we increased the reward density and observed that the navigation policy convergence was quicker and the paths traversed were more efficient.

Finally, in real-world tests, we observed similar crowd navigation capability as in the simulation. However, the physical robot could not move smoothly at the velocities setting used in the simulation.

\begin{figure*}
	\centering
	\includegraphics[width=\textwidth]{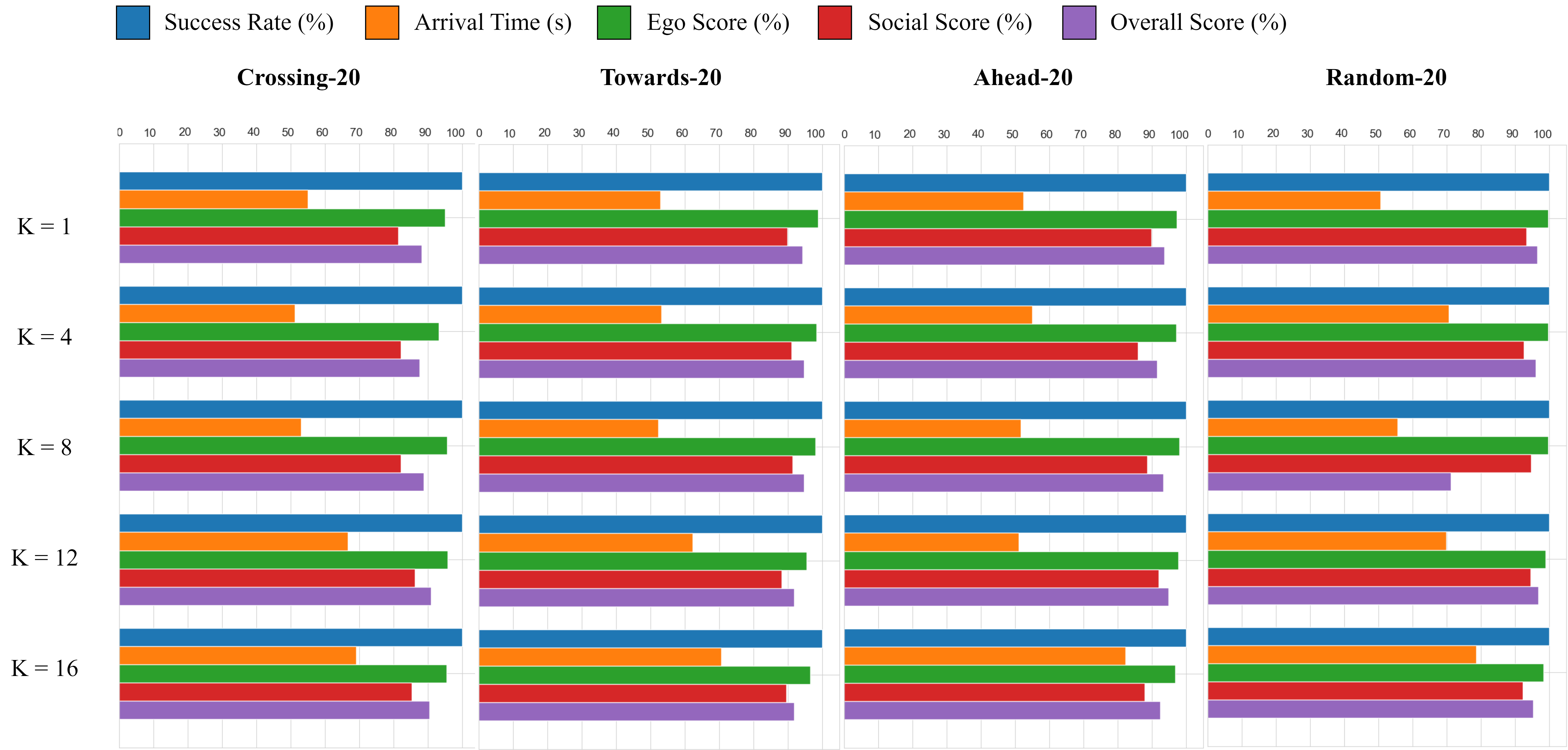}
	\caption{Performance comparison with different $K$ values. E.g, $K=4$ means the robot pays attention to a maximum of four most concerning moving obstacles when navigation through the crowd.}
	\label{fig:results_comparison_k}
\end{figure*}

\subsection{K obstacles}
Fig. \ref{fig:results_comparison_k} shows the result of our investigation of the effect of changing the value of $K$. The robot learned a policy for each $K$ value and tested with the same $K$ value. We observed that the value of $K$ has an impact on the navigation performance of the system. $K$ = 1 was used to compare with the results of \cite{9197148}. Given that in all cases the success rate was 100\%, we compare the arrival time in each case. The performance at $K=1$, $K=4$ and $K=8$ were similar. However, there was a significant increase in arrival time at high $K$ values of 12 and 16. This indicates the usefulness of limiting the attention to a small number of moving obstacles.

\subsection{Ablation Study}
\label{subsection:ablation}

To investigate the effect of the Collision Probability (CP) (\ref{eqn:cp_total}) and its two components, we trained the $K=1$ model with two variations: one with only the $P_{c-ttc}$ (time to collision CP) component (Model-CP-ttc), and one without CP completely (Model-no-CP). The results are shown in the bottom two rows in Fig. \ref{fig:results_comparison}. 

As anticipated, the Model-no-CP and Model-CP-ttc reached the target goal significantly slower than the model with complete risk perception (third row in Fig. \ref{fig:results_comparison}). During the tests, the robot was observed to avoid obstacles altogether by trying to detour. Without CP, the model was not able to estimate collision risk, so it learned that the best way to avoid collision was by avoiding the obstacles completely. Without just CP-ttc, the robot has learned to not take risks and took a longer time to reach the goal. In both models, the robot managed to reach the target goal location. The addition of $P_{c-dto}$ (distance to obstacle CP) has improved the risk estimation as shown in the superior performance of the model with full CP (third row in Fig. \ref{fig:results_comparison}).

\section{CONCLUSIONS}

We have developed a navigation approach for mobile robots using 2D laser scans to improve their performance in crowded environments. Our experiments have shown that the inclusion of the Collision Probability of top $K$ most dangerous moving obstacle to the observation space and local waypoint rewards has achieved improved performance in crowd navigation. Our model was trained in one crowd environment setting and tested on four different crowd environment settings. The perception of risk has enabled the robot to take calculated risks in navigating the crowd while the inclusion of waypoints enabled the robot to reach the target goal location. Besides the good performance in the simulated environment, we have also demonstrated the crowd navigation capability of our model in real-world tests. The robot has shown promising performance although not as dexterous as in the simulation. We plan to expand the real-world tests and improve real-world performance in our future work. We will also investigate further ways to incorporate perceived risk or human awareness in our crowd navigation approach.

The source code and video demonstration of this work are made publicly available on GitHub \cite{hafiqanas2023git}.

\bibliography{root}

\addtolength{\textheight}{-12cm}   





\end{document}